\documentclass[11pt,a4paper]{article}

\usepackage{tacl2021v1}
\taclpubformattrue

\usepackage{times}
\usepackage{latexsym}
\usepackage{multirow}
\usepackage{graphicx}
\usepackage{enumitem}
\usepackage{amsmath}
\usepackage{amssymb}
\usepackage{latexsym}
\usepackage[inkscapelatex=false]{svg}
\usepackage{url}

\usepackage{ulem}   
\usepackage[linesnumbered,ruled,vlined]{algorithm2e} 
\usepackage{subfigure}
\usepackage{makecell}
\usepackage{color}
\usepackage{booktabs}
\usepackage{hyperref}
\usepackage{comment}
\usepackage[all]{nowidow}
\usepackage[subtle]{savetrees} 
\usepackage[T1]{fontenc}
\definecolor{darkred}{RGB}{150, 0, 0}
\definecolor{darkgreen}{RGB}{0, 150, 0}

\usepackage[utf8]{inputenc}

\usepackage{microtype}

\usepackage{inconsolata}

\SetKwInput{KwInput}{Input}
\SetKwInput{KwOutput}{Output}

\usepackage{float}
\usepackage{dsfont}
\newcommand{\graphmoe}{\textsc{GraphMoE}}
\definecolor{shape}{rgb}{0.0,0.5,0.0}

\title{\graphmoe: Amplifying Cognitive Depth of Mixture-of-Experts Network \\via Introducing Self-Rethinking Mechanism}

\author{Bo Lv\textsuperscript{1}\footnotemark[2], Chen Tang\textsuperscript{2}\footnotemark[2], Zifan Zheng\textsuperscript{2,6}\footnotemark[2], Bohao Yang\textsuperscript{3}, Kun Zhao\textsuperscript{4}, Ning Liao\textsuperscript{2}, \\
\textbf{Xiaoxing Wang\textsuperscript{2}}, \textbf{Feiyu Xiong\textsuperscript{2}}, \textbf{Zhiyu Li\textsuperscript{2}\footnotemark[1]}, \textbf{Nayu Liu\textsuperscript{1}\footnotemark[1]} ~and \textbf{Jingchi Jiang\textsuperscript{5}\footnotemark[1]}  \\
  \textsuperscript{1}Institute of Computing Technology, Chinese Academy of Sciences \\
  \textsuperscript{2}Institute for Advanced Algorithms Research, Shanghai \\  
  \textsuperscript{3} The University of Manchester, \textsuperscript{4} The University of Pittsburgh \\
  \textsuperscript{5} Harbin Institute of Technology, \textsuperscript{6} University of Sydney \\
  \texttt{travistang@foxmail.com}}

\begin{document}
\maketitle


\begin{abstract}

Traditional Mixture-of-Experts (MoE) networks benefit from utilizing multiple smaller expert models as opposed to a single large network. However, these experts typically operate independently, leaving a question open about whether interconnecting these models could enhance the performance of MoE networks. In response, we introduce \graphmoe, a novel method aimed at augmenting the cognitive depth of language models via a self-rethinking mechanism constructed on Pseudo Graph MoE networks. \graphmoe{} employs a recurrent routing strategy to simulate iterative thinking steps, thereby facilitating the flow of information among expert nodes.
We implement the \graphmoe{} architecture using Low-Rank Adaptation techniques (LoRA) and conduct extensive experiments on various benchmark datasets. The experimental results reveal that \graphmoe{} outperforms other LoRA based models, achieving state-of-the-art (SOTA) performance. Additionally, this study explores a novel recurrent routing strategy that may inspire further advancements in enhancing the reasoning capabilities of language models. Our code is available at \url{https://github.com/fan2goa1/GraphMoE}
\end{abstract}

\begin{figure}[ht]
\centering
\includegraphics[width=1.0\linewidth]{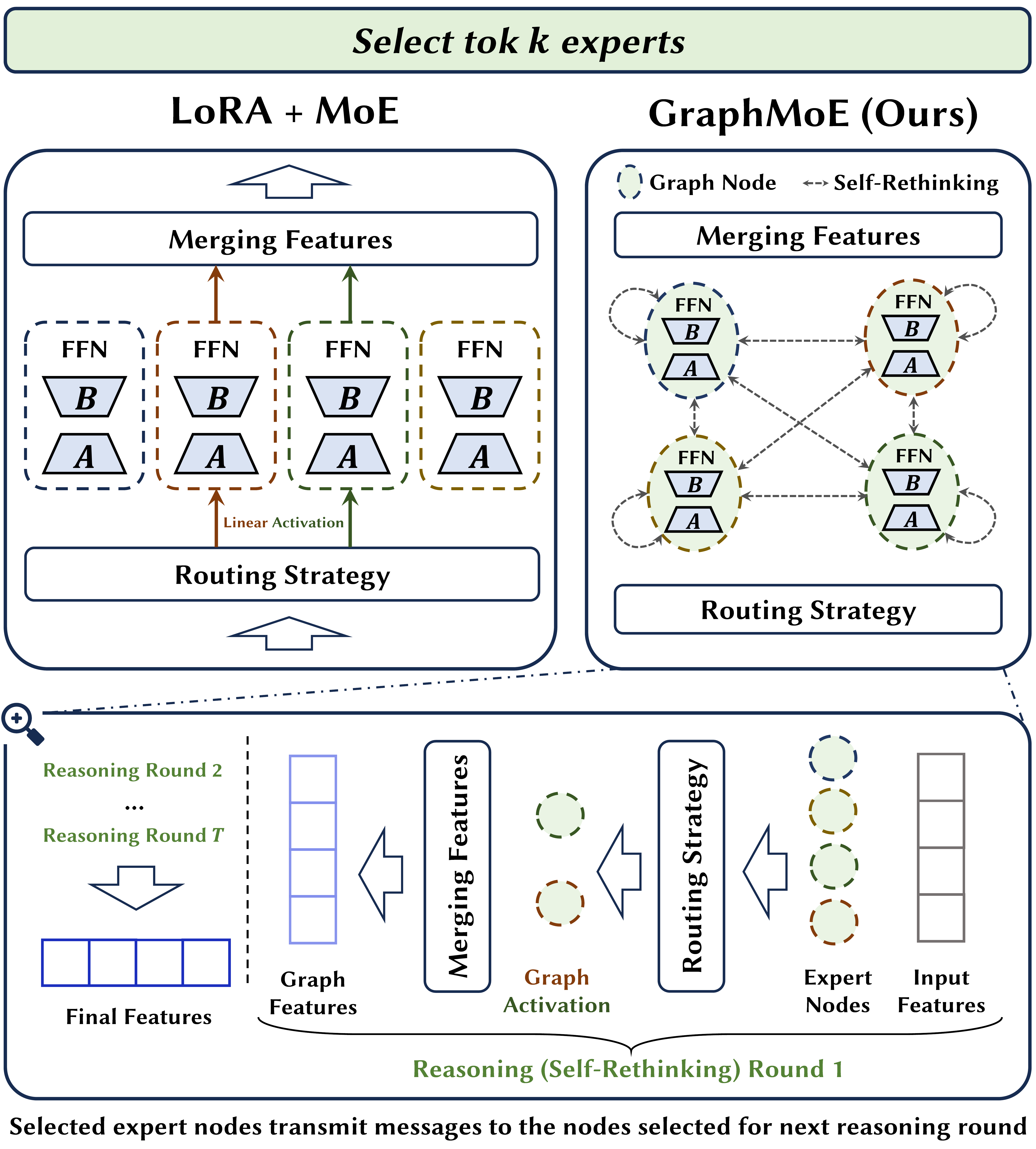}
\vspace{-1em}
\caption{Comparison between LoRA+MoE and \graphmoe{} architectures.}
\label{fig:intro}
\end{figure}

\section{Introduction}
Large Language Models (LLMs), exemplified by the GPT series, have dramatically advanced the field of Natural Language Processing (NLP), demonstrating exceptional performance across a variety of tasks including commonsense reasoning \cite{yang-etal-2024-effective}, creative generation \cite{loakman-etal-2023-twistlist}, dialogue generation \cite{lv-etal-2024-urg, zhao2023evaluating, zhao2024slide}, and summarization \cite{goldsack-etal-2023-enhancing, zhao2024x}. In recent years, there has been a growing interest in refining LLM architectures to deliver superior performance with reduced parameters and lower GPU memory consumption \cite{tang-etal-2023-enhancing}.
A prominent strategy in this endeavor is the Mixture-of-Experts (MoE) architecture \cite{cai2024survey}, which pre-trains multiple Feed-Forward Networks (FFNs) as specialized experts, activating a select few during inference. This strategy allows experts to operate optimally within their domains of expertise, providing enhanced performance compared to LLMs with a similar number of activated parameters.

However, expert models are employed independently through a linear routing strategy. Given that these models are optimized for distinct input distributions, we hypothesize that fostering collaboration among expert models akin to connected nodes in a graph network may further exploit their problem-solving capabilities. Inspired by the knowledge aggregation on pseudo graph~\cite{tang-etal-2023-enhancing}, we propose employing a recurrent routing strategy wherein expert models function as graph nodes. We posit that with increased graphical message transmission between expert nodes, these models can progressively narrow the semantic gap between the MoE model outputs and human references. Thus, we introduce \graphmoe{} (Graph Mixture of Experts), a pioneering approach aimed at deepening the cognitive capabilities of language models by integrating a self-rethinking mechanism into constructed pseudo-graph MoE networks. This methodology may bolster the reasoning capabilities of LLMs by enabling incremental problem-solving, akin to the Chain-of-Thoughts (CoT) strategy.

\graphmoe{} emulates human-like iterative reasoning by allowing the model to continuously revisit and refine its comprehension of input data through multiple reasoning cycles\footnote{This motivation is based on the dual process theory in cognitive science, constructing a pseudo-graph iteratively by integrating a base model for direct inference (System 1) with a LoRA-based rethinking module for deliberate reasoning (System 2). The graph architecture is particularly apt for creating such an iterative thinking system, and related implementations can be found in \citet{ding-etal-2019-cognitive,yan2023complex}.}. This is accomplished by implementing a recurrent routing strategy on a pseudo-graph formed by expert models. These models transmit their outputs as signals for selecting subsequent expert batches and as inputs encapsulating aggregated graph features from prior reasoning rounds—a mechanism we term the ``self-rethinking mechanism''. Given the challenges inherent in pre-training a MoE LLM from scratch, we chose to implement the \graphmoe{} architecture utilizing Low-Rank Adaptation (LoRA) techniques\footnote{As stated in \S\ref{subsec:PEFT}, LoRA introduces a set of additional, small-scale training parameters to fine-tune a LLM for a downstream task, while keeping most of the LLM's original parameters frozen.}, enabling evaluation of \graphmoe{} via Supervised Fine-Tuning (SFT) across several benchmarks.
 
In \autoref{fig:intro}, we present a comprehensive illustration of \graphmoe{}(LoRA) and conduct a comparative analysis with the traditional LoRA+MoE framework. In the conventional LoRA+MoE approach, the features of activated expert models are integrated directly. This process renders other inactive expert models non-contributory to problem-solving, particularly when the model tends to consistently select a specific set of expert models, as evidenced by low workload balancing. In contrast, \graphmoe{}(LoRA) conceptualizes expert nodes as graph nodes, enabling message transmission across these interconnected nodes through graph activation over multiple reasoning rounds. This approach maintains the reduced memory usage characteristic of conventional LoRA+MoE models (with same activation number of expert models in each cycle) while increasing computational iterations. Our experimental results confirm the effectiveness of \graphmoe, demonstrating its superior performance compared to other LoRA baselines and achieving state-of-the-art (SOTA) results.

The main contributions of this work are as follows:
\begin{itemize}
    \item To the best of our knowledge, this research represents the first attempt to introduce a ``self-rethinking mechanism'', aiming to enable neural networks to emulate a human-like iterative reasoning process. This mechanism enhances the capability of MoE networks to engage in complex cognitive functions.
    
    \item We propose a novel pseudo-graph-based MoE architecture designed to facilitate the flow of information between expert nodes, thereby improving representation learning and enhancing the performance of LLMs.
    
    \item Comprehensive experiments have been conducted to validate the effectiveness of the proposed method. These experiments highlight the method's efficacy in augmenting the cognitive depth of language models.
    
    \item We investigate the potential of the ``self-rethinking mechanism'' in improving LLMs and identify key factors that contribute to enhancing their reasoning abilities, offering valuable insights for future research.
\end{itemize}

\section{Related Work} \label{sec:related_works}

\subsection{Parameter-Efficient Fine-Tuning (PEFT)} \label{subsec:PEFT}
LLMs have demonstrated remarkable improvements in general natural language understanding (NLU) and natural language generation (NLG) tasks \citep{guo2023evaluating,chang2024survey,tang2024knowledge}. However, their performance often suffers from limited generalization and effectiveness in domain-specific applications. To address this, numerous studies have proposed Parameter-Efficient Fine-Tuning (PEFT) methods, which optimize only a small subset of LLM parameters, significantly reducing computational overhead.

One of the most popular PEFT methods is Low-Rank Adaptation (LoRA) \citep{LoRA_21_NIPS_Google}, which introduces two low-rank matrices to each weight matrix $\mathbf{W}$ in LLMs. The product of these matrices represents the weight adjustment $\Delta \mathbf{W}$. Building upon LoRA, subsequent works have introduced more efficient variants, such as VeRa \citep{VeRa_2023_arXiv_UAmsterdam}, AdaLoRA \citep{AdaLoRA_23_ICLR_GIT}, DoRA \citep{DoRA_24_arXiv_NVIDIA}, and MoSLoRA \citep{wu2024mixture}. These methods aim to better capture task-specific features and integrate diverse feature subspaces effectively.

\subsection{Transformer \& Recurrent Models}
The self-attention mechanism, as a fundamental algorithm within Transformers \citep{Transformer_17_NIPS_Google}, facilitates parallel sequence processing and enhances model capacity through increased width. However, it lacks the temporal reasoning capabilities inherent in recurrent architectures, which contribute to model depth, such as Long Short-Term Memory (LSTM) networks \citep{hochreiter1997long} and Gated Recurrent Units (GRUs) \citep{GRU_14_arXiv_Montr}. To address this limitation, recent research has investigated integrating Transformer architectures with recurrent structures \citep{RWKV_23_arXiv_EleutherAI,RecurLinearTrans_23_arXiv_UAlberta,CachedTrans_24_AAAI_CUHK,tang2024knowledge}. These efforts aim to blend recurrent neural network capabilities with the ability to capture long-distance dependencies. In this study, we likewise endeavor to incorporate a recurrent mechanism to model multiple expert systems. However, rather than focusing on capturing long-distance dependencies, our primary objective is to emulate the stepwise cognitive processes characteristic of human cognition. To achieve this, we employ GRUs to augment reasoning depth by aggregating hidden representations from attention features at each stage of the recurrent routing process. This integration is designed to enhance the model's proficiency in apprehending complex dependencies and thereby improve its overall reasoning capabilities.

\subsection{Mixture of Experts (MoE)} \label{subsec:MoE}
The concept of Mixture of Experts (MoE) was first introduced by \citet{jacobs1991adaptive}, who proposed training multiple networks (experts) on different subsets of data and aggregating their outputs. Recently, as LLMs have become a focal point of research, MoE layers have been integrated into Transformer-based architectures. Specifically, researchers have replaced standard Feed-Forward Networks (FFNs) with sparse MoE layers, employing novel routing strategies \citep{zuo2021taming,zhong2024lory,wu2024yuan,muqeeth2023soft,fu2024moa} and advanced expert segmentation techniques \citep{dai2024deepseekmoe,he2024mixture,jiang2024mixtral,xiao2024improving}.

Numerous studies have investigated the application of Parameter-Efficient Fine-Tuning (PEFT) methods to introduce additional trainable parameters for implementing pseudo MoE structures within LLMs \citep{dou-etal-2024-loramoe,luo2024moelora,gao2024higher,li2024mixlora,gou2023mixture}. These models expand the conventional single feed-forward network (FFN) architecture and its corresponding representation space into multiple subspaces, thereby effectively emulating the behavior of MoE architectures. 

Prominent examples include MoLA \citep{gao2024higher}, LoRAMoE \citep{dou-etal-2024-loramoe}, and MixLoRA \citep{li2024mixlora}, all of which have demonstrated state-of-the-art performance compared to other PEFT-based methods across various benchmarks. Consequently, we integrate these three distinct LoRA MoE methodologies into our \graphmoe{} framework. The primary differences between these models are as follows: LoRAMoE incorporates vanilla attention LoRA layers and MLP plugged with LoRA-MoE layers; MoLA integrates plugged LoRA-MoE layers in both attention and MLP layers; and MixLoRA combines fused LoRA-MoE modules in attention and MLP layers. Further analysis of these LoRA-MoE implementations can be found in \citet{li2024mixlora}.

In this study, we augment MoE architectures by enhancing their reasoning depth and demonstrate the efficacy of our novel architecture when integrated with PEFT-based LLM baselines.

\begin{figure*}[htbp]
\centering
\includegraphics[width=1.0\linewidth]{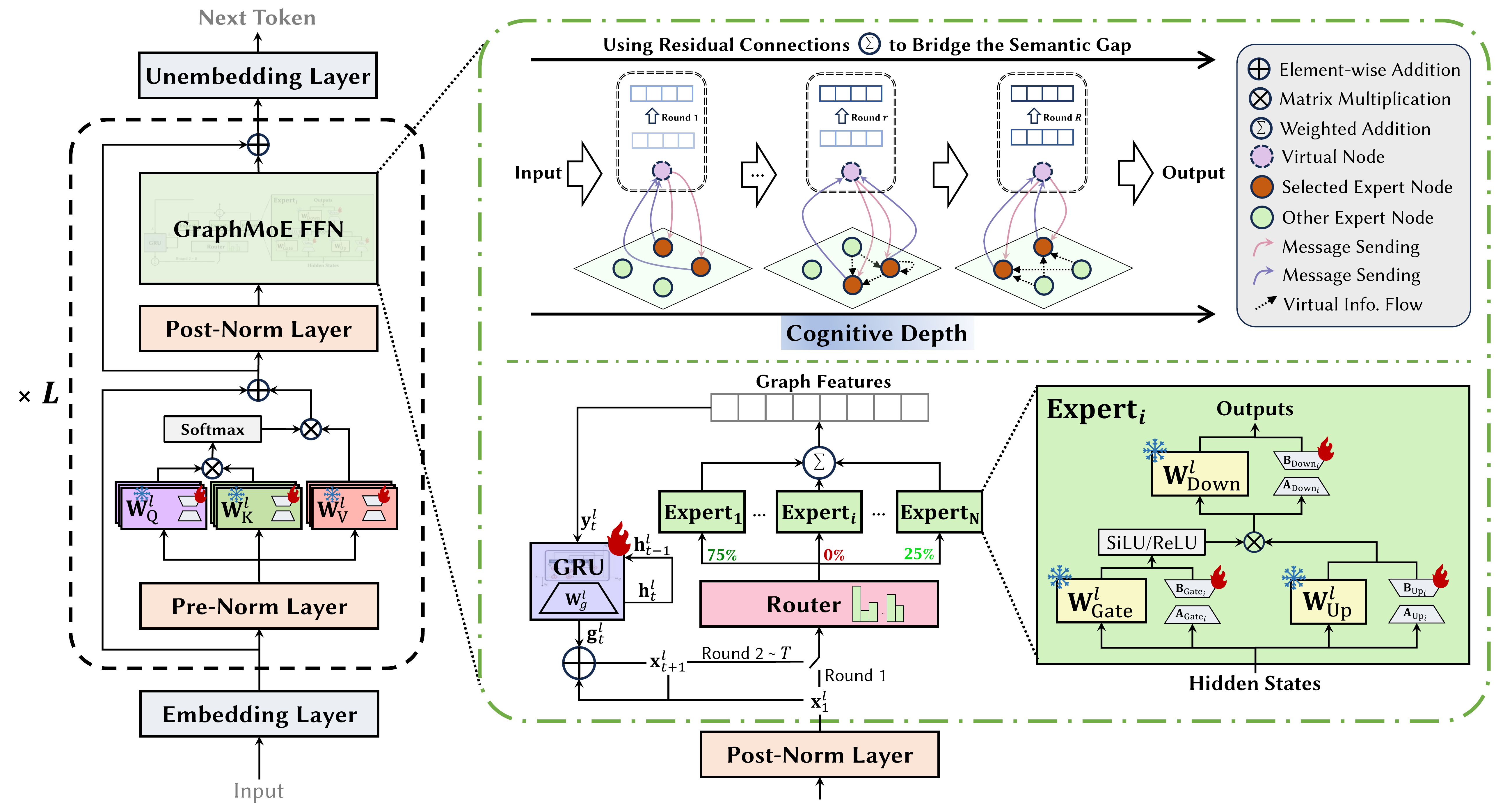}
\vspace{-2em}
\caption{Overview of \graphmoe{} architecture. In this figure, the original Feed-Forward Network (FFN) layer in each transformer block is modified. FFN is also known as a Multi-Layer Perceptron (MLP).}
\label{fig:method}
\end{figure*}

\section{Method}
In this section, we present the construction of the pseudo-graph network of MoE model designed to interconnect expert models as graph nodes, facilitating information exchange among these models in a manner akin to the human brain. In this analogy, different neurons are responsible for specific abilities and collaborate through signal transmission across synapses. We apply our \graphmoe{} methodology to additional LoRA networks, and introduce our approach in three distinct subsections: \S\ref{subsec:moe_transform} MoE Transformation, \S\ref{subsec:pseudo_graph} Pseudo Reasoning Graph Construction, and \S\ref{subsec:graphmoe_training} \graphmoe{} Training.

The overall framework is illustrated in \autoref{fig:method}. The self-rethinking mechanism operates within the GraphMoE Feed-Forward Network (FFN) layer, which replaces the original FFN layer found in standard transformer-based LLMs. In this diagram, the LoRA MoE is derived from the MixLoRA architecture. When using alternative architectures such as MoLA and LoRAMoE, the illustration of the Expert component varies. Nonetheless, the implementation of the self-rethinking module remains consistent, as it incorporates an additional GRU model introduced in \S\ref{subsec:pseudo_graph}.

\subsection{MoE Transformation} \label{subsec:moe_transform}
Incorporating multiple expert models into LoRA layers, inspired by the work of \citet{li2024mixlora} and \citet{dou-etal-2024-loramoe}, the standard Feed-Forward Networks (FFNs) can be converted into MoE structures comprising $n$ experts, denoted as $\{E_i\}_{i=1}^n$. As illustrated in the bottom-right corner of Figure~\ref{fig:method}, for the $\ell$-th layer of the LLM $\left(1 \leq \ell \leq L\right)$, we freeze the original FFN parameters and train three pairs of low-rank matrices for each expert: $\mathbf{A}_{\text{Down}_i}^{\ell} \in \mathbb{R}^{r \times d_{1}}$, $\mathbf{B}_{\text{Down}_i}^{\ell} \in \mathbb{R}^{d_{2} \times r}$, $\mathbf{A}_{\text{Gate}_i}^{\ell}$, $\mathbf{B}_{\text{Gate}_i}^{\ell}$, $\mathbf{A}_{\text{Up}_i}^{\ell}$, and $\mathbf{B}_{\text{Up}_i}^{\ell}$. Here, $i$ represents the $i$-th expert, and $d_1$, $d_2$ are the dimensions of the corresponding weight matrices $\mathbf{W}_{\text{Down}_i}^\ell \in \mathbb{R}^{d_2 \times d_1}$, with other dimensions defined similarly. Each expert $E_i(\cdot)$ is constructed as a fusion of the original FFN and trainable low-rank matrices, as shown in Equations~\ref{equ:moe_expert} and \ref{equ:expert_lora}, where $\sigma(\cdot)$ denotes activation functions (e.g., SiLU, ReLU), and $\odot$ denotes element-wise multiplication.
\begin{equation} \label{equ:moe_expert}
E^{\ell}_{i}(\mathbf{x}) = \mathbf{\tilde{W}}_{\text{Down}_i}^\ell \left(\sigma (\mathbf{\tilde{W}}_{\text{Gate}_i}^\ell \mathbf{x}) \odot (\mathbf{\tilde{W}}_{\text{Up}_i}^\ell \mathbf{x}) \right)
\end{equation}
\begin{equation} \label{equ:expert_lora}
\begin{aligned}
    & \mathbf{\tilde{W}}_{\text{Gate}_i}^\ell = \mathbf{W}_{\text{Gate}_i}^\ell + \mathbf{B}_{\text{Gate}_i}^\ell \cdot \mathbf{A}_{\text{Gate}_i}^\ell \\
    & \mathbf{\tilde{W}}_{\text{Down}_i}^\ell = \mathbf{W}_{\text{Down}_i}^\ell + \mathbf{B}_{\text{Down}_i}^\ell \cdot \mathbf{A}_{\text{Down}_i}^\ell \\
    & \mathbf{\tilde{W}}_{\text{Up}_i}^\ell = \mathbf{W}_{\text{Up}_i}^\ell + \mathbf{B}_{\text{Up}_i}^\ell \cdot \mathbf{A}_{\text{Up}_i}^\ell
\end{aligned}
\end{equation}

Additionally, we introduce a trainable linear layer $\mathbf{R}^{\ell} \in \mathbb{R}^{n \times d}$, referred to as the Router, where $n$ denotes the total number of experts, and $d$ represents the dimension of the hidden state. The Router determines the importance of each expert during forward propagation by computing a relevance score for every expert. Based on the Router's output, we select the top $k$ experts and normalize their corresponding weights to ensure effective routing, as described in Equations~\ref{equ:router} and \ref{equ:topk}. In subsequent experiments, unless stated otherwise, our setup activates 2 out of 8 experts, thus setting $k$ to 2.

\begin{equation} \label{equ:router}
\mathbf{\hat{s}}^\ell = \operatorname{Softmax}\left(\mathbf{R}^{\ell} \cdot \mathbf{x}\right)
\end{equation}
\begin{equation} \label{equ:topk}
\begin{aligned}
    \mathbf{s}^\ell &= \operatorname{Top-}k \left( \mathbf{\hat{s}}^\ell \right) \\
    &= 
    \begin{cases} 
        \mathbf{\hat{s}}^\ell\left[i\right], & \text{if } \mathbf{\hat{s}}^\ell\left[i\right] \text{ is in the top } k, \\
        0, & \text{otherwise.}
    \end{cases}
\end{aligned}
\end{equation}

Finally, as shown in Equation~\ref{equ:weighted_sum}, the output of the MoE module is computed as a weighted sum of the selected experts. This result replaces the original output of the vanilla FFN module.
\begin{equation} \label{equ:weighted_sum}
    \mathbf{y}^{\ell} = MoE\left(\mathbf{x}\right) = \sum_{i=1}^{n}\ \mathbf{s}^\ell[i] \cdot E_i^{\ell}(\mathbf{x})
\end{equation}

\subsection{Pseudo Reasoning Graph Construction} \label{subsec:pseudo_graph}

The development of the pseudo reasoning graph represents a pivotal step in implementing our proposed self-rethinking mechanism, which is designed to enhance the internal consistency of LLMs \citep{liang2024internal}. The pseudocode outlining the inference process of \graphmoe{} is detailed in Algorithm~\ref{alg:graphmoe}. For clarity and conciseness, the terms Gate, Up, and Down in the subscripts are abbreviated as $G$, $U$, and $D$, respectively.

Initially, the pseudo graph is constructed as illustrated in the top right corner of \autoref{fig:method}. Within this framework, expert models are represented as graph nodes, each functioning according to the procedure described in \S\ref{subsec:moe_transform}. A distinctive virtual node is simultaneously established alongside these expert model graph nodes. Unlike other nodes, the virtual node is characterized by a graph feature vector and a GRU module. This node serves as a recurrent router, acting as a conduit for expert node connectivity and a layer for forward feature aggregation. 

In contrast to conventional knowledge graphs, there is no prior knowledge that defines the edges between nodes. Instead, the edges connecting graph nodes are dynamically generated at each reasoning round, denoted as $t$, with the virtual node serving as the intermediary, as indicated by the black dashed arrows in Figure~\ref{fig:method}. This process facilitates the dynamical activation of temporal edges between nodes, enabling feature transformation across nodes and the newly created edges. Consequently, expert nodes at each reasoning round can communicate and collaborate to address the given task effectively.

\begin{figure}[htbp]
\centering
\includegraphics[width=1.0\linewidth]{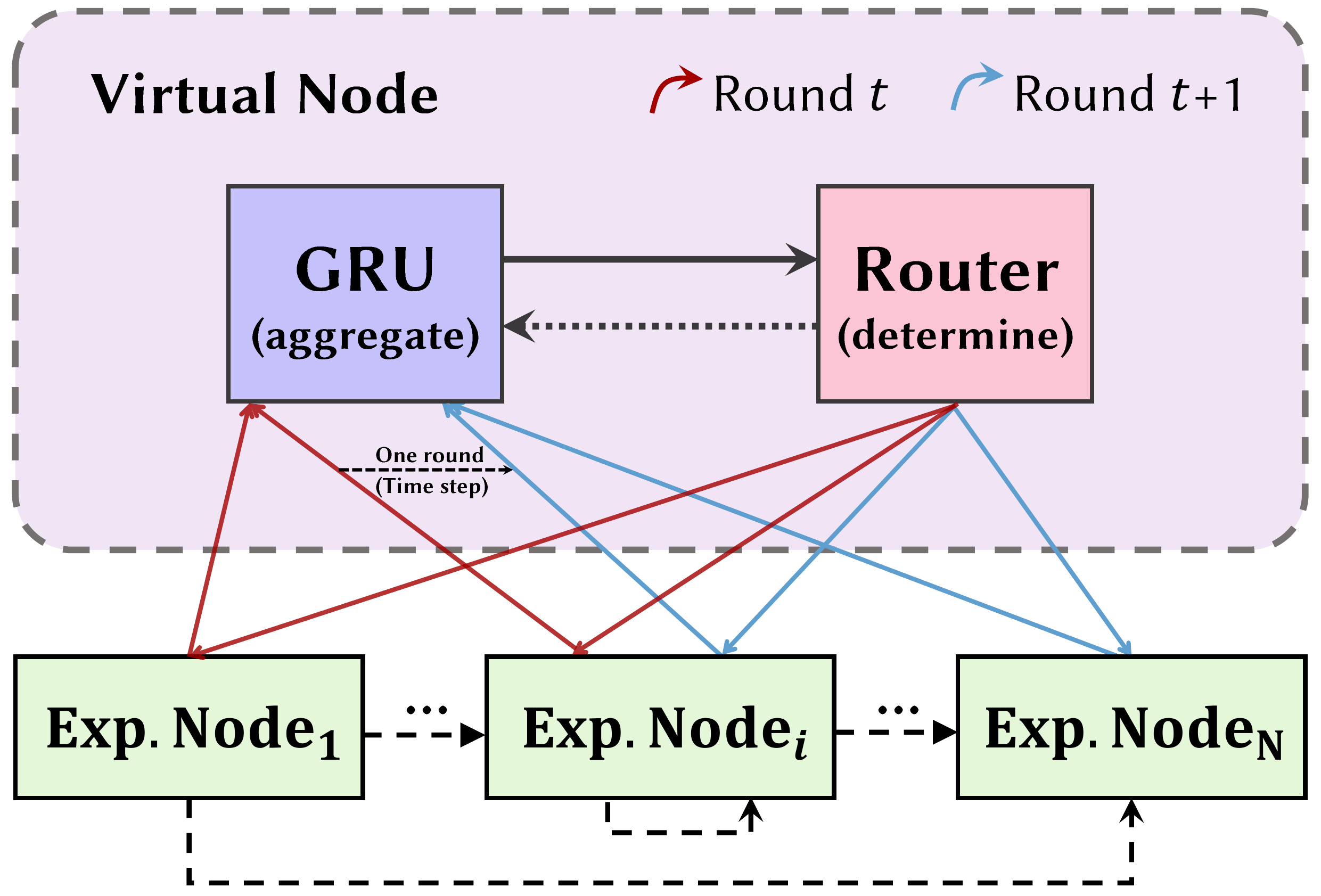}
\vspace{-2em}
\caption{The mechanism how a virtual node collect and aggregate features from expert graph nodles.}
\label{fig:virtual_node}
\end{figure}

Therefore, the core of this method lies in implementing the functionality of the virtual node. In this work, we employ a Low-Rank Gated Recurrent Unit\footnote{Due to space constraints, we do not elaborate on the structure of the standard GRU \citep{GRU_14_arXiv_Montr} here.} (GRU) to serve as the virtual node. Specifically, we reduce the GRU's hidden size to $\overline{d} \ll d$ and incorporate a low-rank linear layer to compute its output $\mathbf{g}^{\ell}_{t}$. The virtual node integrates features from previous reasoning round and processes the expert results $\mathbf{y}^{\ell}$, subsequently selecting $k$ experts for the next round.

The reasoning (self-thinking) process involves a total of $T$ rounds of MoE computations. The first round is processed as described in Equations~\ref{equ:moe_expert}-\ref{equ:weighted_sum}. After obtaining the representation $\mathbf{y}_{t}^{\ell}$ from the MoE, it is fed into the Low-Rank GRU along with the hidden state $\mathbf{h}_{t-1}^{\ell}$ from the previous timestep (here, “round” and “timestep” are equivalent since there is only one step per round). A residual connection is applied by adding the result to the input of the previous round, $\mathbf{x}_{t}^{\ell}$, to produce the input for the next round, $\mathbf{x}_{t+1}^{\ell}$. This iterative process extracts meta-information for subsequent reasoning rounds, as shown in Equations~\ref{equ:gru}-\ref{equ:residual}. 

\begin{equation} \label{equ:gru}
\begin{aligned}
\mathbf{z}_t^{\ell} =\sigma&\left(\mathbf{W}^{\ell}_z \cdot\left[\mathbf{h}^{\ell}_{t-1}, \mathbf{y}^{\ell}_t\right]\right) \text{, } \mathbf{W}^{\ell}_z \in \mathbb{R}^{\overline{d} \times (\overline{d}+d)} \\
\mathbf{r}^{\ell}_t =\sigma&\left(\mathbf{W}^{\ell}_r \cdot\left[\mathbf{h}^{\ell}_{t-1}, \mathbf{y}^{\ell}_t\right]\right) \text{, } \mathbf{W}^{\ell}_r \in \mathbb{R}^{\overline{d} \times (\overline{d}+d)} \\
\mathbf{\hat{h}}^{\ell}_t & =\sigma \left(\mathbf{W}^{\ell}_o \cdot\left[\mathbf{r}_t^{\ell} \odot \mathbf{h}_{t-1}^{\ell}, \mathbf{\mathbf{y}}_t^{\ell}\right]+\mathbf{b}_{o}^{\ell}\right) \\
\mathbf{h}^{\ell}_t & =\left(\mathbf{1}-\mathbf{z}_t^{\ell}\right) \odot \mathbf{h}^{\ell}_{t-1}+\mathbf{z}_t^{\ell} \odot \mathbf{\hat{h}}_t^{\ell}
\end{aligned}
\end{equation}

\begin{equation} \label{equ:residual}
\begin{aligned}
\mathbf{g}^{\ell}_{t} = \mathbf{W}_g^{\ell} &\cdot \mathbf{h}^{\ell}_t \text{, } \mathbf{W}_g^{\ell} \in \mathbb{R}^{d \times \overline{d}} \\
\mathbf{x}^{\ell}_{t+1} &= \mathbf{x}^{\ell}_t + \mathbf{g}^{\ell}_{t}
\end{aligned}
\end{equation}

In a nutshell, from a model architecture standpoint, the GRU and the Router together form the aforementioned ``virtual node''. Specifically, the GRU functions to aggregate information from multiple experts $E_{i}$, whereas the Router is responsible for determining which experts should receive the integrated information for further inference, as depicted in \autoref{fig:virtual_node}.

\begin{algorithm}[ht] \label{alg:graphmoe}
\caption{\graphmoe{} at Layer $\ell$}
\KwInput{$\mathbf{x}_{1}^{\ell}$: \textcolor{shape}{$(\mathtt{B}, \mathtt{P}, \mathtt{D})$} after the Post-Norm.}
\KwOutput{$\mathbf{y}^{\ell}_{T}$: \textcolor{shape}{$(\mathtt{B}, \mathtt{P}, \mathtt{D})$} the hidden states.}
\textcolor{gray}{/* \textcolor{shape}{$\mathtt{B}$} for \text{Batch Num,} \textcolor{shape}{$\mathtt{P}$} for Seq Len, \textcolor{shape}{$\mathtt{D}$} for Dim */}

\textcolor{gray}{/* \textcolor{shape}{$\mathtt{D'}$} for \text{Upper Dim,} \textcolor{shape}{$\overline{\mathtt{D}}$} for GRU Hidden Dim */}

\For{$t \gets 1$ \textbf{to} $T$}{
    \textcolor{gray}{\text{/* MoE Forward */}}
    
    $\mathbf{\tilde{s}}^{\ell}_{t}$: \textcolor{shape}{$(\mathtt{B}, \mathtt{P}, \mathtt{N})$} $\gets$ $\operatorname{Softmax} \left(\mathbf{R}^{\ell} * \mathbf{x}^{\ell}_{t}\right)$\;
    
    $\mathbf{s}^{\ell}_{t}$: \textcolor{shape}{$(\mathtt{B}, \mathtt{P}, \mathtt{N})$} $\gets$ $\operatorname{Top-}k\left(\mathbf{\tilde{s}}^{\ell}_{1}\right)$\;
    
    \For{$i$ in $\left\{\text{selected experts } E\right\}$}{
        $\tilde{\mathbf{W}}_{\text{G}_i}^{\ell}$: \textcolor{shape}{$(\mathtt{D'}, \mathtt{D})$} $\gets$ $\mathbf{W}_{\text{G}_i}^\ell + \mathbf{B}_{\text{G}_i}^\ell * \mathbf{A}_{\text{G}_i}^\ell$\;



        \text{Similarly, we can get } $\tilde{\mathbf{W}}_{\text{U}_i}^{\ell}$ \text{and} $\tilde{\mathbf{W}}_{\text{D}_i}^{\ell}$\;

        $\hat{\mathbf{c}}^\text{G}_{i}$: \textcolor{shape}{$(\mathtt{B}, \mathtt{P}, \mathtt{D'})$} $\gets$ $\tilde{\mathbf{W}}_{\text{G}_i}^{\ell} * \mathbf{x}_{t}^{\ell}$\;

        $\hat{\mathbf{c}}^\text{U}_{i}$: \textcolor{shape}{$(\mathtt{B}, \mathtt{P}, \mathtt{D'})$} $\gets$ $\tilde{\mathbf{W}}_{\text{U}_i}^{\ell} * \mathbf{x}_{t}^{\ell}$\;

        $\hat{\mathbf{c}_{i}}$: \textcolor{shape}{$(\mathtt{B}, \mathtt{P}, \mathtt{D'})$} $\gets$ $\sigma\left(\hat{\mathbf{c}}^\text{G}_{i}\right) \odot \hat{\mathbf{c}}^\text{U}_{i} $\;

        $\mathbf{c}_{i}$: \textcolor{shape}{$(\mathtt{B}, \mathtt{P}, \mathtt{D})$} $\gets$ $\tilde{\mathbf{W}}_{\text{D}_i}^{\ell} * \hat{\mathbf{c}_{i}}$\;

         $\mathbf{y}^{\ell}_{t}$: \textcolor{shape}{$(\mathtt{B}, \mathtt{P}, \mathtt{D})$} $\gets$ $\mathbf{y}^{\ell}_{t} + \mathbf{s}^{\ell}_{t}\left[:,i\right] \odot \mathbf{c}_{i}$\;
    }

    \textcolor{gray}{\text{/* Low-Rank GRU */}}
    
    \If{$t \ne T$}{
        $\mathbf{z}_t^{\ell}$: \textcolor{shape}{$(\mathtt{B}, \mathtt{P}, \overline{\mathtt{D}})$} $\gets$ $\sigma\left(\mathbf{W}^{\ell}_z * \left[\mathbf{h}^{\ell}_{t-1}, \mathbf{y}^{\ell}_{t}\right]\right)$\;
        
        $\mathbf{r}_t^{\ell}$: \textcolor{shape}{$(\mathtt{B}, \mathtt{P}, \overline{\mathtt{D}})$} $\gets$ $\sigma\left(\mathbf{W}^{\ell}_r * \left[\mathbf{h}^{\ell}_{t-1}, \mathbf{y}^{\ell}_{t}\right]\right)$\;

        $\mathbf{\hat{h}}^{\ell}_t$: \textcolor{shape}{$(\mathtt{B}, \mathtt{P}, \overline{\mathtt{D}}+\mathtt{D})$} $\gets$ $\left[\mathbf{r}_t^{\ell} \odot \mathbf{h}_{t-1}^{\ell}, \mathbf{\mathbf{x}}_t^{\ell}\right]$\;

        $\mathbf{h}^{\ell}_t$: \textcolor{shape}{$(\mathtt{B}, \mathtt{P}, \overline{\mathtt{D}})$} $\gets$ $\sigma \left(\mathbf{W}^{\ell}_o * \mathbf{\hat{h}}^{\ell}_t+\mathbf{b}_{o}^{\ell}\right)$\;

        $\mathbf{h}^{\ell}_t$ $\gets$ $\left(\mathbf{1}-\mathbf{z}_t^{\ell}\right) \odot \mathbf{h}^{\ell}_{t-1}+\mathbf{z}_t^{\ell} \odot \mathbf{h}_t^{\ell}$\;

        $\mathbf{g}^{\ell}_{t}$: \textcolor{shape}{$(\mathtt{B}, \mathtt{P}, \mathtt{D})$} $\gets$  $\mathbf{W}_g^{\ell} \cdot \mathbf{h}^{\ell}_t$\;

        \textcolor{gray}{\text{/* Residual Connection */}}
        
        $\mathbf{x}^{\ell}_{t+1}$: \textcolor{shape}{$(\mathtt{B}, \mathtt{P}, \mathtt{D})$} $\gets$ $\mathbf{x}^{\ell}_t + \mathbf{g}^{\ell}_{t}$
    }
   
}

\Return{$\mathbf{y}^{\ell}_{T}$: \textcolor{shape}{$(\mathtt{B}, \mathtt{P}, \mathtt{D})$}}\;
\end{algorithm}

To enhance clarity in illustrating the long processing pipeline, we summarize the entire methodology in Algorithm \ref{alg:graphmoe}. The process involves $T$ reasoning iterations, referred to as self-rethinking. Initially, features from the FFN layer of transformer blocks are derived via a LoRA MoE forward pass. Subsequently, these features are transmitted to the subsequent Low-Rank GRU module, which computes both the residual connection within the pseudo graph and the merged feature for the input of the succeeding iteration. 

\subsection{\graphmoe{} Training} \label{subsec:graphmoe_training}
Building upon the architecture introduced in \S\ref{subsec:moe_transform} and \S\ref{subsec:pseudo_graph}, this section provides a detailed explanation of the training methodology for \graphmoe{}.

\paragraph{Trainable Components.} As shown in Figure~\ref{fig:method}, the modules marked with a flame icon represent the trainable components in \graphmoe{}. Specifically, we need to train $3 \cdot L \cdot n$ pairs of low-rank adapters applied to the MoE modules, as well as a single linear layer serving as the Router. Additionally, in each decoder layer, the Low-Rank GRU module requires training 4 linear layers: $\mathbf{W}^{\ell}_z$, $\mathbf{W}^{\ell}_r$, $\mathbf{W}^{\ell}_o$, and $\mathbf{W}^{\ell}_g$. 

Inspired by \citet{li2024mixlora}, we also enhance the attention module by adding 4 pairs of trainable low-rank matrices to each decoder layer. These matrices are applied to the attention components $\mathbf{Q}^{\ell}_{h}$, $\mathbf{K}^{\ell}_{h}$, $\mathbf{V}^{\ell}_{h} \in \mathbb{R}^{d \times \frac{d}{H}}$, and $\mathbf{O}^{\ell}_{h} \in \mathbb{R}^{\frac{d}{H} \times d}$, where $h$ and $H$ represent the $h$-th attention head and the total number of heads, respectively. This enhances the adaptability of certain attention heads to specific tasks \citep{zheng2024attention}.

In summary, as shown in Table~\ref{tab:TrainComponent}, \graphmoe{} training involves a total of $(3n + 4) \cdot L$ pairs of low-rank adapters and $4 \cdot L$ linear layers.
\begin{table}[htbp]
\centering
\resizebox{\linewidth}{!}{%
\begin{tabular}{@{}ccccc@{}}
\toprule
\multicolumn{1}{c}{\textbf{Component}} & \textbf{Attention Head} & \textbf{MoE}      & \textbf{GRU} & \textbf{Total}  \\ \midrule
Low-rank adapters                      & $4\cdot L$              & $3\cdot L\cdot n$ & 0            & $(3n+4)\cdot L$ \\
Linear layers                          & 0                       & 0                 & $4\cdot L$   & $4\cdot L$      \\ \bottomrule
\end{tabular}%
}
\vspace{-0.7em}
\caption{Summary of components to be trained.}
\label{tab:TrainComponent}
\end{table}

\paragraph{Loss Design.} Similar to standard continual learning~\cite{wu2024continual} and supervised fine-tuning~\cite{patil2024review}, the primary objective during training is to minimize the cross-entropy loss $\mathcal{L}_{CE}$ between the predicted token logits and the target tokens.

However, the Router is a unique and critical component of the MoE architecture, as it determines which experts are activated during inference. Prior research has shown that unconstrained Router training can lead to over-reliance on a few specific experts \citep{STMoE_22_arXiv_Google,SwiTrans_22_JMLR_Google}, resulting in poor load balancing. Inspired by the work of \citet{JetMoE_24_arXiv_MIT} and \citet{OLMoE_24_arXiv_Allen}, we introduce a \textbf{Load Balancing Loss} $\mathcal{L}_{LB}$ (as defined in Equation~\ref{equ:loss_lb}) as an auxiliary objective. This loss penalizes the Router for disproportionately selecting the same experts across the sequence set $\mathcal{M}$, which consists of several sequences and the corresponding next tokens to be predicted.
\begin{equation} \label{equ:loss_lb}
\begin{aligned}
\mathcal{L}_{LB} =& n\cdot \sum_{i=1}^{n} f_i \cdot p_i\text{,}\\
\text{where } f_i &= \frac{1}{|\mathcal{M}|} \sum_{\left(\mathbf{x}, \mathbf{y}\right)\in\mathcal{M}} \mathds{1}\{\mathbf{s}_{\mathbf{x}}\left[i\right] > 0\}\text{,}\\
p_i &= \frac{1}{|\mathcal{M}|} \sum_{\left(\mathbf{x}, \mathbf{y}\right)\in\mathcal{M}} \mathbf{s}_{\mathbf{x}}\left[i\right]
\end{aligned}
\end{equation}
Setting a hyperparameter $\lambda$, the total loss can be expressed as follows:
\begin{equation} \label{equ:total_loss}
\mathcal{L}_{\text {total}}=\mathcal{L}_{CE}+\lambda \cdot \mathcal{L}_{LB}
\end{equation}


\section{Experiment}

\subsection{Experimental Settings}
As there are multiple methodologies for applying MoE to LoRA layers, 
the aforementioned \graphmoe{} architecture is developed on the foundation of existing LoRA combined with MoE base models, designated as \graphmoe{}(base model). In the \graphmoe{}(base model), the conventional MoE module is substituted with our proposed pseudo reasoning graph (PRG).

To evaluate \graphmoe{}, we select a diverse range of commonsense reasoning datasets that offer unique challenges across multiple task types. These include question-answer tasks such as ARC \cite{clark2018think}, OpenBookQA \cite{mihaylov2018can}, PIQA \cite{bisk2020piqa}, and SocialIQA \cite{sap2019socialiqa}, each designed to assess different aspects of commonsense and contextual reasoning. For classification, we use BoolQ \cite{clark2019boolq}, which tests the model's ability to handle yes/no questions. Hellaswag \cite{zellers2019hellaswag} is employed for science completion tasks, requiring selection of the most plausible outcomes from rich contextual setups. Winogrande \cite{sakaguchi2021winogrande} addresses fill-in-the-blank tasks, facilitating evaluation of nuanced language understanding through large-scale ambiguities. 

These datasets collectively provide a platform for assessing our model's accuracy and generalization across various reasoning aspects. Experiments are conducted using brain floating point 16-bit (BF16), as using full precise float 32 (FP32) results in experiments being over four times slower, which is unacceptable given our computing constraints.

\subsection{Baselines}
In our experimental setup, all baselines are evaluated by fine-tuning LLMs on datasets. This is achieved by keeping the parameters of the base language model frozen and training only the additional adapters. We use the LLaMA-3-8B model as our base language model throughout the experiments. To benchmark our proposed \graphmoe{} model, we chose to compare it against the traditional Low-Rank Adaptation (LoRA)~\cite{hu2021lora} approach, as well as three recently introduced SOTA LoRA+MoE methods: MoLA~\cite{gao2024higher}, LoRAMoE~\cite{dou-etal-2024-loramoe}, and MixLoRA~\cite{li2024mixlora}. Considering 
Weight-Decomposed Low-Rank Adaptation (DoRA)~\cite{DoRA_24_arXiv_NVIDIA} has been shown to effectively enhance the LoRA method, we also develop DoRA variants for each of these selected baseline models to benchmark against our method.

\subsection{Evaluation Metrics}
The evaluation of all methods was conducted using the accuracy metric across all datasets. In order to accurately extract the answers provided by LLMs for Multiple-Choice Question-Answer (MCQA) tasks, we adopt the first-token evaluation approach, which captures the probabilities of each letter choice \citep{yu2024xfinder}. In assessing the workload balance of expert models, we recorded the frequency of selection for each expert model during the evaluation phase and calculated their proportions of the total selections. For further insight, we introduced several indicating variables such as GRU hidden size scale ($\overline{d}/d$), parameter size (Param), and inference time (Infer Time). The GRU hidden size scale is determined by dividing the GRU hidden size by the dimension of the LLM model. Parameter size indicates the ratio of trainable parameters to the total number of parameters. Inference Time is defined as the ratio of inference time to the minimum value in a series of inference times.

\subsection{Implementation Details}
Our hyperparameters strictly follow the configurations detailed in \cite{li2024mixlora} including the LoRA and DoRA methods and their three MoE derivatives, e.g. MixLoRA or MixDoRA. We summarize the settings for LoRA/DoRA and their derivatives of MoE-based methods, as shown in Table \ref{tab:train_settings}. During evaluation, the batch size is set to $8$.


\begin{table}[htbp]
\centering
\label{tab:train_settings}
\resizebox{\columnwidth}{!}{%
\begin{tabular}{@{}c|ccc@{}}
\toprule
\multicolumn{1}{c|}{\textbf{Setting}} & \textbf{LoRA/DoRA}                                                                                                                   & \textbf{MixLoRA/MixDoRA}                                                                                                            & \textbf{\graphmoe{}}                                                                                                                                                                                                                                                                                                                              \\ \midrule
Cutoff Length                         & \multicolumn{3}{c}{512}                                                                                                                                                                                                                                                                                                                                                                                                                                                                                                                                                                                                        \\
Learning Rate                         & \multicolumn{3}{c}{2e-4}                                                                                                                                                                                                                                                                                                                                                                                                                                                                                                                                                                                                       \\
Optimizer                             & \multicolumn{3}{c}{AdamW}                                                                                                                                                                                                                                                                                                                                                                                                                                                                                                                                                                                                      \\
Batch Size                            & \multicolumn{3}{c}{16}                                                                                                                                                                                                                                                                                                                                                                                                                                                                                                                                                                                                         \\
Accumulation Steps                    & \multicolumn{3}{c}{8}                                                                                                                                                                                                                                                                                                                                                                                                                                                                                                                                                                                                          \\
Dropout Rate                          & \multicolumn{3}{c}{0.05}                                                                                                                                                                                                                                                                                                                                                                                                                                                                                                                                                                                                       \\
Training Epochs                       & \multicolumn{3}{c}{2}                                                                                                                                                                                                                                                                                                                                                                                                                                                                                                                                                                                                          \\ \midrule
LoRA Rank ($r$)                       & \multicolumn{1}{c|}{80}                                                                                                              & \multicolumn{2}{c}{16}                                                                                                                                                                                                                                                                                                                                                                                                                                                                  \\
LoRA Alpha ($\alpha$)                 & \multicolumn{1}{c|}{160}                                                                                                             & \multicolumn{2}{c}{32}                                                                                                                                                                                                                                                                                                                                                                                                                                                                  \\
Expert Number                            & \multicolumn{1}{c|}{-}                                                                                                               & \multicolumn{2}{c}{8}                                                                                                                                                                                                                                                                                                                                                                                                                                                                   \\
Top-$k$                               & \multicolumn{1}{c|}{-}                                                                                                               & \multicolumn{2}{c}{2}                                                                                                                                                                                                                                                                                                                                                                                                                                                                   \\ \midrule
Targeted Parameters                   & \multicolumn{2}{c|}{\begin{tabular}[c]{@{}c@{}}$\mathbf{Q}^{\ell}_{h}$, $\mathbf{K}^{\ell}_{h}$, $\mathbf{V}^{\ell}_{h}$,  $\mathbf{O}^{\ell}_{h}$, \\ $\mathbf{W}_{\text{Gate}_i}^\ell$, $\mathbf{W}_{\text{Up}_i}^\ell$, $\mathbf{W}_{\text{Down}_i}^\ell$\end{tabular}} & \begin{tabular}[c]{@{}c@{}}$\mathbf{Q}^{\ell}_{h}$, $\mathbf{K}^{\ell}_{h}$, $\mathbf{V}^{\ell}_{h}$, $\mathbf{O}^{\ell}_{h}$,\\ $\mathbf{W}_{\text{Gate}_i}^\ell$, $\mathbf{W}_{\text{Up}_i}^\ell$, $\mathbf{W}_{\text{Down}_i}^\ell$,\\ $\mathbf{W}^{\ell}_z$, $\mathbf{W}^{\ell}_r$, $\mathbf{W}^{\ell}_o$, $\mathbf{W}^{\ell}_g$\end{tabular} \\ \bottomrule
\end{tabular}%
}
\vspace{-0.7em}
\caption{Training Settings for baselines and \graphmoe{}.}
\end{table}

The experiments described in this paper required a single A800 GPU for at least one month. Including debugging and downtime, the entire process extended to around two months. The base model, LLaMA-8B, was obtained from the Huggingface repository ``meta-llama/Meta-Llama-3-8B-Instruct''\footnote{\url{https://huggingface.co/meta-llama/Meta-Llama-3-8B-Instruct}}. 

In the following experiments utilizing the \graphmoe{} framework, a range of main hyperparameters are configured as follows: (1) The number of iterations for applying the GRU gate within the recurrent routing, referred to as the reasoning (self-rethinking) round $T$; our \graphmoe{} models are configured with a default setting of $3$. During the initial round, the outputs of expert models are obtained without the GRU gate's feature transmission. (2) The GRU network is set to a single layer. (3) The GRU hidden size scale is predefined as $0.1 \times 4096$\footnote{The actual hidden size is approximated to the nearest integer using Python fucntion ``int()''.}, where $4096$ represents the model dimension of the LLaMA-3-8B model.
(4) The hyperparameter \(\lambda\) associated with the auxiliary load balance training loss \(\mathcal{L}_{LB}\) is configured to be $0.01$. (5) Unless otherwise specified, the expert activation in the MoE architecture is set to a total of 8 experts per MoE module, with 2 experts activated each time.\footnote{Additional discussions regarding the implementation details of this paper can be found in \autoref{apx:discussion}.}

\section{Experimental Result}

\subsection{Evaluating the Performance of \graphmoe{}} 

\begin{table*}[htbp]
\centering
\resizebox{0.9\linewidth}{!}{
\begin{tabular}{l|cccccccc|c}
    \toprule
    \textbf{Method}   & \textbf{ARC-E} & \textbf{ARC-C} & \textbf{BoolQ} & \textbf{OBQA} & \textbf{PIQA} & \textbf{SIQA} & \textbf{HellaS} & \textbf{WinoG} & \textbf{AVG.} \\
    \midrule
    MoLA  & 86.4 & 77.9 & 74.0 & 84.4 & 86.7 & 76.4 & 93.9 & 83.3 & 82.9 \\
    \graphmoe{} (MoLA) & 89.7 & 81.0 & 75.8 & 87.6 & 87.8 & 79.7 & 95.5 & 83.2 & \textbf{85.0} \\
    $\Delta$ & \textcolor{darkgreen}{3.3} & \textcolor{darkgreen}{3.1} & \textcolor{darkgreen}{1.8} & \textcolor{darkgreen}{3.2} & \textcolor{darkgreen}{1.1} & \textcolor{darkgreen}{3.3} & \textcolor{darkgreen}{1.6} & \textcolor{darkred}{-0.1} & \textcolor{darkgreen}{2.1} \\
    
    \midrule
    
    LoRAMoE  & 87.8 & 79.5 & 72.4 & 85.0 & 87.1 & 74.8 & 94.8 & 83.4 & 83.1 \\
     \graphmoe{} (LoRAMoE) & 87.7 & 79.9 & 75.9 & 88.8 & 87.8 & 79.4 & 94.8 & 83.4 & \textbf{84.7} \\
     $\Delta$ & \textcolor{darkred}{-0.1} & \textcolor{darkgreen}{0.4} & \textcolor{darkgreen}{3.5} & \textcolor{darkgreen}{3.8} & \textcolor{darkgreen}{0.7} & \textcolor{darkgreen}{4.6} & 0.0 & 0.0 & \textcolor{darkgreen}{1.6} \\
     
    \midrule
    
    MixLoRA & 86.9 & 77.0 & 74.0 & 84.4 & 86.0 & 75.5 & 93.7 & 83.3 & 82.6\\
    \graphmoe{} (MixLoRA) & 89.2 & 80.9 & 75.8 & 89.6 & 87.8 & 77.0 & 95.4 & 83.2 & \textbf{84.9} \\
    $\Delta$ & \textcolor{darkgreen}{2.3} & \textcolor{darkgreen}{3.9} & \textcolor{darkgreen}{1.8} & \textcolor{darkgreen}{5.2} & \textcolor{darkgreen}{1.8} & \textcolor{darkgreen}{1.5} & \textcolor{darkgreen}{1.7} & \textcolor{darkred}{-0.1} & \textcolor{darkgreen}{2.3} \\
    \bottomrule
\end{tabular}
}
\caption{Comparison of the performance between baseline LoRA+MoE models and \graphmoe{} (LoRA+MoE) models. \graphmoe{} ($\cdot$) indicates that the conventional MoE module in the baseline has been substituted with the self-rethinking mechanism proposed in \S~\ref{subsec:pseudo_graph}. Scores highlighted in \textbf{bold} denote superior performance for each respective method. \textcolor{darkgreen}{Green} and \textcolor{darkred}{red} indicate increase and decrease in performance following the application of \graphmoe{} respectively.}
\label{tab:transferability}
\end{table*}
As illustrated in \autoref{tab:transferability}, the implementation of \graphmoe{}, a pseudo graph-based MoE network incorporating a recurrent routing strategy, has resulted in a significant performance enhancement across all SOTA LoRA+MoE baselines. This underscores \graphmoe{}'s ability to boost performance by fostering improved collaboration among expert models, achieved solely by replacing the MoE module within different methods. The observed improvement stems from an optimized mechanism utilizing the expert models to learn the representations with the feature transmission among themselves.

The degree of enhancements varies across models, influenced by their specific base architectures. This variability is attributed to the integration level, or the combination mode, between the MoE and LoRA components, which can impact the parameter space for incremental learning and thereby modify the marginal benefits of \graphmoe{}. Specifically, as highlighted by \citet{li2024mixlora}, the MoLA and LoRAMoE configurations incorporate LoRA modules as expert models by integrating them with outputs from either the attention layers or MLP layers—an approach termed the ``Plugged-in'' method. In contrast, MixLoRA employs a ``fused'' integration strategy, embedding the LoRA expert models directly within the original network layers. This design allows the LoRA experts to exert a more immediate and substantial influence on the conventional LLM components, resulting in the most significant enhancement of the original LoRA+MoE architecture.

As observed in \autoref{tab:transferability}, \graphmoe{} consistently exerts positive effects—or at least avoids significant negative impacts—on model accuracy across diverse tasks. In some instances, only one out of eight tasks shows a slight accuracy decrease of $0.1$, and these instances occur with different tasks when applying \graphmoe{} to the three baselines\footnote{This can be attributed to experimental error.}. This suggests that \graphmoe{} is robust and reliable in enhancing the performance of MoE architectures. Furthermore, when examining its impact on individual tasks, \graphmoe{} can lead to remarkable improvements; for instance, when applied to LoRAMoE, it boosts the performance on \textbf{SIQA} from $74.8$ to $79.4$, representing an increase of $4.6$ (over $6\%$), which is particularly impressive for this task. This observation further underscores the effectiveness of the \graphmoe{} approach within this architectural framework.

\subsection{Main Results} 

\begin{table*}[htbp]
\centering
\resizebox{0.95\linewidth}{!}{
\begin{tabular}{l|c|cccccccc|c}
    \toprule
    \textbf{Method} & \textbf{Param}  & \textbf{ARC-E} & \textbf{ARC-C} & \textbf{BoolQ} & \textbf{OBQA} & \textbf{PIQA} & \textbf{SIQA} & \textbf{HellaS} & \textbf{WinoG} & \textbf{AVG.} \\
    \midrule
    Base (ICL-$3$) & 0\% & 90.6 &	76.7 & 45.9 & 73.8 & 61.6 & 71.5 & 64.1 & 59.2 & 65.6 \\
    Base (ICL-$3$, Self-cons-$5$) & 0\% & 91.1 & 79.4 & 50.2 & 73.7 & 57.0 & 70.3 & 66.9 & 62.1 & 66.3 \\
    Base (ICL-$3$, Self-cons-$10$) & 0\% & 91.6 & 78.1 & 42.8 & 75.3 & 59.8 & 71.3 & 69.9 & 61.4 & 66.2 \\
    Base (ICL-$3$, Self-cons-$15$) & 0\% & 91.9 & 78.0 & 45.6 & 72.0 & 67.9 & 71.7 & 63.3 & 63.2 & 66.4 \\
    \midrule
    LoRA* & 2.6\% & 89.0 & 75.7 & 67.2 & 85.0 & 80.7 & 78.3 & 74.2 & 75.3 & 78.2 \\
    DoRA* & 2.6\% & 88.1 & 76.4 & 61.7 & 80.6 & 82.3 & 76.2 & 78.8 & 83.7 & 78.5 \\
    MixLoRA* & 3.0\% & 86.5 & 79.9 & 75.0 & 84.8 & 87.6 & 78.8 & 93.3 & 82.1 & 83.5 \\
    MixDoRA* & 3.0\% & 87.7 & 78.9 & 76.8 & 86.9 & 83.4 & 80.1 & 94.6 & 84.2 & 84.1 \\
    \midrule
    LoRA & 2.6\% & 86.2 & 76.7 & 70.2 & 78.2 & 82.9 & 74.2 & 80.1 & 50.7 & 74.9 \\
    DoRA & 2.6\% & 87.6 & 76.8 & 69.3 & 80.8 & 84.2 & 76.8 & 85.3 & 50.4 & 76.4 \\
    MoLA & 2.7\% & 86.4 & 77.9 & 74.0 & 84.4 & 86.7 & 76.4 & 93.9 & 83.3 & 82.9 \\
    MoDA & 2.7\% & 86.4 & 78.0 & 74.0 & 84.8 & 87.4 & 76.4 & \textbf{95.5} & \textbf{84.1} & 83.3 \\
    LoRAMoE & 3.2\% & 87.8 & 79.5 & 72.4 & 85.0 & 87.1 & 74.8 & 94.8 & 83.4 & 83.1 \\
    DoRAMoE & 3.2\% & 87.9 & 80.1 & 74.6 & 85.8 & 86.3 & 74.5 & 94.0 & 83.1 & 83.3\\
    MixLoRA & 3.0\% & 86.9 & 77.0 & 74.0 & 84.4 & 86.0 & 75.5 & 93.7 & 83.3 & 82.6\\
    MixDoRA & 3.0\% & 86.7 & 76.7 & 75.5 & 84.8 & 85.2 & 76.9 & 93.0 & 83.1 & 82.7 \\
    \midrule
    \graphmoe{}(MixLoRA) & 5.9\% & \underline{89.2} & \textbf{80.9} & \underline{75.8} & \textbf{89.6} & \underline{87.8} & \underline{77.0} & \underline{95.4} & 83.2 & \underline{84.9} \\
    \graphmoe{}(MixDoRA) & 5.9\% & \textbf{90.3} & \underline{80.6} & \textbf{75.9} & \underline{88.2} & \textbf{88.8} & \textbf{79.4} & 95.3 & \underline{83.7} & \textbf{85.3} \\
    \bottomrule
\end{tabular}
}
\caption{Performance evaluation of LoRA-Based method SFT across various benchmarks. The term ``ICL-n'' indicates that n examples are included in the prompt for in-context learning, while ``Self-cons-5'' means that 5 samples are collected for self-consistency. ``Param'' refers to the size of the trainable parameters used in the experiments. Results marked with an asterisk (*) are experimental data sourced from \citet{li2024mixlora}, conducted with full precision (FP32). All other results were obtained using BF16 precision. In the table, derivatives of baselines that employ the DoRA technique are indicated by replacing ``L'' with ``D'' in the original model name.}
\label{tab:main}
\end{table*}

The experimental results are detailed in \autoref{tab:main}. The initial part of the experiment involves four base language models (LLaMA-3-8B) as reference models that incorporate in-context learning and self-consistency techniques. In-context learning (ICL-n) refers to directly providing the task's instructions in the prompt along with a certain number of demonstration examples. Self-consistency (Self-cons-n) refers to the common technique in LLMs where multiple reasoning paths are sampled and the most consistent answer is selected. Increasing the samples for self-consistency to 5, 10, and 15 yields performance scores of 66.3, 66.2, and 66.4, respectively, indicating minimal variation. Although self-consistency results in slight performance improvements beyond 65.6 on downstream tasks, the base language model still falls short when compared to those employing the LoRA-based SFT method.

Due to limitations in computational resources, the experiments conducted in this study employ reduced precision BF16 for both training and inference phases. However, we include the full precision results of MixLoRA as a reference for comparison. It is evident that all reproduced models experience a significant performance decline compared to their full precision counterparts. Nonetheless, our  \graphmoe{} models with MixLoRA and MixDoRA as base models still achieve SOTA performance even compared to the full precision results. This substantial improvements compared to other SOTA LoRA+MoE models further demonstrate the effectiveness of the \graphmoe{} architecture. 

Moreover, MixLoRA does not represent the optimal base model with the LoRA+MoE framework, as MoLA and LoRAMoE have better accuracy scores. This indicates that the performance gains achieved by \graphmoe{} surpass those of existing methodologies applied within the LoRA+MoE framework. The introduction of a self-rethinking mechanism is crucial in enhancing the representation learning capabilities inherent to the MoE approach and potentially even language models more broadly. Furthermore, it can be observed that the enhancements afforded by our method are applicable to both LoRA and DoRA. This suggests that the improvements carried by the PRG are not merely a result of simple adjustments on additional trainable parameters, but rather arise from the effective collaboration among diverse expert models facilitated by our approach. This conclusion supports the assertion that the proposed \graphmoe{} framework is indeed effective and the self-rethinking mechanism holds great potential.

\subsection{Analysis of Workload Balance}

\begin{figure*}[tb]
\centering
\includegraphics[width=1.0\linewidth]{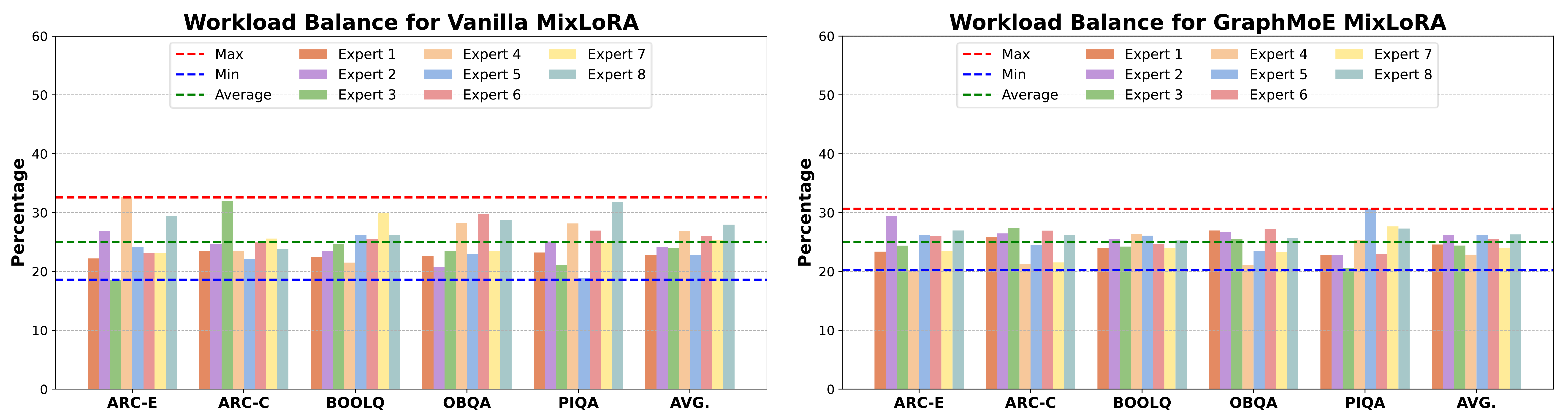}
\vspace{-2em}
\caption{The workloads of all experts are shown by normalizing the selected time during every routing step. \textcolor{red}{Red} and \textcolor{blue}{blue} dashed lines indicate the maximum and minimum workloads across all tasks, while the \textcolor{darkgreen}{green} dashed line shows the average workload. The MixLoRA model and the \graphmoe{} model have standard deviations of $0.0313$ and $0.0215$, respectively, in workload balance.}
\label{fig:workload}
\end{figure*}

The primary distinction between MixLoRA and \graphmoe{}(MixLoRA) lies in the self-rethinking mechanism facilitated by the recurrent router.
We conducted a further analysis of the impact of expert model selection at each routing stage. \autoref{fig:workload} presents a comparative analysis of workload distribution among expert models within the frameworks of MixLoRA and \graphmoe{}(MixLoRA) across various tasks, highlighting the distribution patterns of expert selection. The findings indicate that the \graphmoe{} routing strategy enables a more balanced selection of expert models, as demonstrated by the standard deviation metrics of $0.0313$ for MixLoRA and $0.0249$ for \graphmoe{}(MixLoRA), representing a reduction of over $20\%$ in standard deviations across $8$ experts.

The aforementioned phenomenon suggests that multiple iterations of the reasoning process allow for a greater number of experts to be involved in addressing problems, thereby enhancing the effectiveness of MoE in representation learning. In each reasoning step, a distinct set of expert models is more likely to be activated to incrementally construct the representations within the MoE. This approach allows more expert models to operate in their optimal conditions, sequentially addressing complex problems. It is as if a chain-of-thoughts (CoT) is implicitly formed, using different expert models to fit features at different stages.

\subsection{Sensitivity Analysis and Overhead}

\begin{figure*}[tb]
\centering
\includegraphics[width=1.0\linewidth]{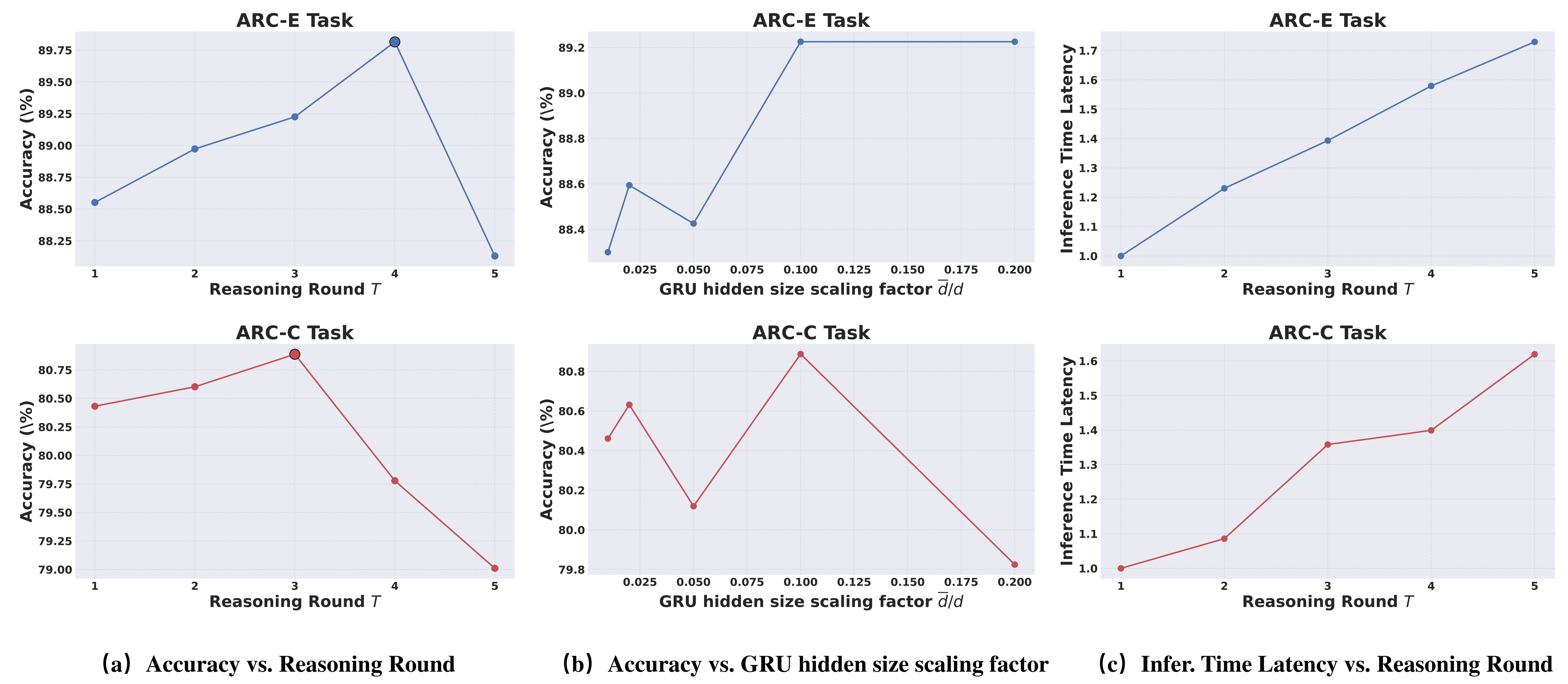}
\vspace{-0.7em}
\caption{Sensitivity analysis of the additional hyperparameters in the \graphmoe{} architecture. The impact on \graphmoe{}'s computational overhead is demonstrated by examining how the inference time scales with the Reasoning Round ($T$). In subfigure (c), ``Infer. Time'' refers to the inference time, with the duration of the first round set as a unit. The figure displays the factors by which the time of each subsequent round compares to that of the first round.}
\label{fig:sensitivity}
\end{figure*}

Due to computational constraints, our sensitivity analysis was conducted exclusively on the ARC-E and ARC-C tasks, as these have the smallest dataset sizes. This choice allowed for significantly faster model execution compared to other tasks. The results are depicted in Figure \ref{fig:sensitivity}. It is evident that the hyperparameters selected in previous experiments were not optimal, suggesting that \graphmoe{} could achieve greater improvements in base models with a better selection of hyperparameters. The primary objective of this sensitivity analysis is to provide insights into the efficacy of our self-rethinking mechanism.

In subfigure (a) of Figure \ref{fig:sensitivity}, as the Reasoning Round $T$ increases, both tasks exhibit an increase in accuracy, indicating that the self-rethinking mechanism indeed enhances the effectiveness of MoE by deepening its cognitive processing. 
However, accuracy significantly declines once $T$ reaches a certain threshold, specifically $T=4$ for ARC-E and $T=3$ for ARC-C. We postulate that this decline is due to overfitting, implying that the model risks ``overthinking'' the problem.
This is further evidenced by the ARC-C task in subfigure (b) of Figure \ref{fig:sensitivity}, where an increase in the GRU hidden size does not guarantee improved performance. This phenomenon underscores the importance of the self-rethinking mechanism itself over the newly introduced trainable parameters of the GRU. In subfigure (c) of Figure \ref{fig:sensitivity}, it is observed that, compared to conventional LoRA+MoE (MixLoRA in this experiment) models, which employ a single reasoning round, the increase in inference time for \graphmoe{} is only approximately $0.16$ times for each additional reasoning round. This increase is both manageable and acceptable. In summary, these findings demonstrate the efficacy of our proposed method and its promising impact on enhancing the cognitive depth of MoE.

\subsection{Effect of Parameter Size \\ on Model Performance}
\label{apx:param_size}

\begin{table*}[ht]
    \centering
    \resizebox{0.9\linewidth}{!}{
    \begin{tabular}{l|c|ccccccccc}
        \toprule
        \textbf{Method} &\textbf{Param} &\textbf{ARC-E} & \textbf{ARC-C} & \textbf{BoolQ} & \textbf{OBQA} & \textbf{PIQA} & \textbf{SIQA} & \textbf{HellaS} & \textbf{WinoG} & \text{AVG}\\
        \midrule
        LoRA &1.9\% & 87.5 & 78.9 & 69.8 & 84.4 & 84.4 & 77.1 & 88.4 & 49.1 & 77.5\\
        LoRA &2.6\% & 86.2 & 76.7 & 70.2 & 78.2 & 82.9 & 74.2 & 80.1 & 50.7 & 74.9\\
        LoRA &5.9\% & 84.7 & 78.6 & 63.6 & 82.2 & 82.1 & 32.9 & 84.1 & 50.4 & 69.8\\
        LoRA &7.9\% & 83.9 & 73.1 & 62.2 & 84.4 & 82.5 & 74.6 & 92.7 & 49.5 & 75.4\\
        LoRA &9.9\% & 85.9 & 73.3 & 70.0 & 82.0 & 85.7 & 32.1 & 89.1 & 49.6 & 71.0\\
        \midrule
        MixLoRA & 3.0\% & 86.9 & 77.0 & 74.0 & 84.4 & 86.0 & 75.5 & 93.7 & 83.3 & 82.6\\
        MixLoRA &6.0\% & 87.3 & 79.0 & 74.9 & 88.4 & 88.1 & 76.7 & 93.6 & 82.5 & 83.8 \\
        MixLoRA & 9.0\%	& 86.7 & 77.0 & 71.7 & 86.2 & 80.5 & 74.0 &	89.5 & 80.4 & 80.8 \\
        \midrule
         \graphmoe{} & 5.9\% & \textbf{89.2} & \textbf{80.9} & \textbf{75.8} & \textbf{89.6} & \textbf{87.8} & \textbf{77.0} & \textbf{95.4} & \textbf{83.2} & \textbf{84.9}\\
        \bottomrule
    \end{tabular}
    }
    \caption{Performance Comparison of Parameter Variants. In the table, \graphmoe{} refers to the \graphmoe{}(MixLoRA) model.}
    \label{tab:lora_performance}
\end{table*}

It has been claimed in previous works \cite{tang2023terminology, li2024mixlora} that simply increasing parameter size will not give a metric increase when doing SFT. We conduct an experiment to further demonstrate this phenomenon. From \autoref{tab:lora_performance}, it can be observed that increasing the parameter size does not consistently lead to improved performance. For instance, the LoRA method with 1.9\% parameters achieves the best average score of 77.5, while increasing the parameter size to 7.9\% results in an series of fluctuation below the score of 77.5. The similar phenomenon has been also observed in (b) of \autoref{fig:sensitivity}, where the performance is not increasing when the GRU hidden size scaling factor increases. 

Moreover, we directly increase one of the baseline model of \graphmoe{} to show if the increased performance was achieved by the additionally increased parameter size. Even though the performance of MixLoRA with 6.0\% parameters results in a higher average score of 83.9, the larger size of Mixlora with 9.0\% parameters result in an inferior performance. This indicates that the self-rethinking mechanism plays the pivotal role in effective parameter utilization and model structuring, as showcased by the comparison between MixLoRA and \graphmoe{}(MixLoRA), and implementing the \graphmoe{} framework can yield better results without proportionally increasing parameter size.

Overall, it can be observed that the best overall performance is observed with \graphmoe{}(MixLoRA), which uses just 5.9\% of parameters yet achieves the highest average score of 84.9. This demonstrates that optimal model design principles, rather than mere parameter scaling, are crucial for achieving superior results in these tasks.

\begin{table*}[ht]
    \centering
    \resizebox{0.9\linewidth}{!}{
    \begin{tabular}{l|c|cccccccc|c}
        \toprule
        \textbf{Method} &\textbf{Param} &\textbf{ARC-E} & \textbf{ARC-C} & \textbf{BoolQ} & \textbf{OBQA} & \textbf{PIQA} & \textbf{SIQA} & \textbf{HellaS} & \textbf{WinoG} & \text{AVG}\\
        \midrule
        MixLoRA$\dagger$ &  6.0\% & 87.4 & 79.2 & 73.9 & 85.2 & 85.3 & 72.9 &	95.0 & 80.4 & 82.4 \\
        MixLoRA & 6.0\% & 87.3 & 79.0 & 74.9 & 88.4 & 88.1 & 76.7 & 93.6 & 82.5 & \textbf{83.8} \\
        $\Delta$  & - & \textcolor{darkred}{-0.1} & \textcolor{darkred}{-0.2} & \textcolor{darkgreen}{1} & \textcolor{darkgreen}{3.2} & \textcolor{darkgreen}{2.8} & \textcolor{darkgreen}{3.8} & \textcolor{darkred}{-1.4} & \textcolor{darkgreen}{2.1} & \textcolor{darkgreen}{1.4} \\
        \midrule
        \graphmoe{}$\dagger$ & 5.9\% & 88.1 & 79.0 & 74.1 &	87.8 &	85.6 & 74.3 & 93.2 & 85.1 & 83.4 	\\
         \graphmoe{} & 5.9\% & 89.2 & 80.9 & 75.8 & 89.6 & 87.8 & 77.0 & 95.4 & 83.2 & \textbf{84.9}\\
         $\Delta$  & - &  \textcolor{darkgreen}{1.1} & \textcolor{darkgreen}{1.9} & \textcolor{darkgreen}{1.7} & \textcolor{darkgreen}{1.8} & \textcolor{darkgreen}{2.2} & \textcolor{darkgreen}{2.7} & \textcolor{darkgreen}{2.2} & \textcolor{darkred}{-1.9} & \textcolor{darkgreen}{1.5}	\\
        \bottomrule
    \end{tabular}
    }
    \caption{Performance Comparison of \graphmoe{}(MixLoRA) with different activated experts. \graphmoe{}$\dagger$ indicates the \graphmoe{}(MixLoRA) variant with $4$ out of $8$ experts activated, whereas the other LoRA MoE models in above experiments have $2$ out of $8$ experts activated.}
    \label{tab:lora_performance_expert}
\end{table*}

To provide further insights into the utilization of the MoE structure, we experimented with activating additional experts within the MoE layers to assess if similar performance levels could be achieved, and the results are presented in \autoref{tab:lora_performance_expert}. Surprisingly, increasing the number of active experts led to a decline in performance. This finding underscores that additional parameters do not necessarily guarantee improved reasoning performance. Instead, the effectiveness of the designed mechanisms within the neural architecture plays a more critical role, as demonstrated by previous results in \autoref{tab:lora_performance}.

\subsection{In-depth Analysis of Computation Cost}
\begin{figure}[ht]
\centering
\includegraphics[width=1.0\linewidth]{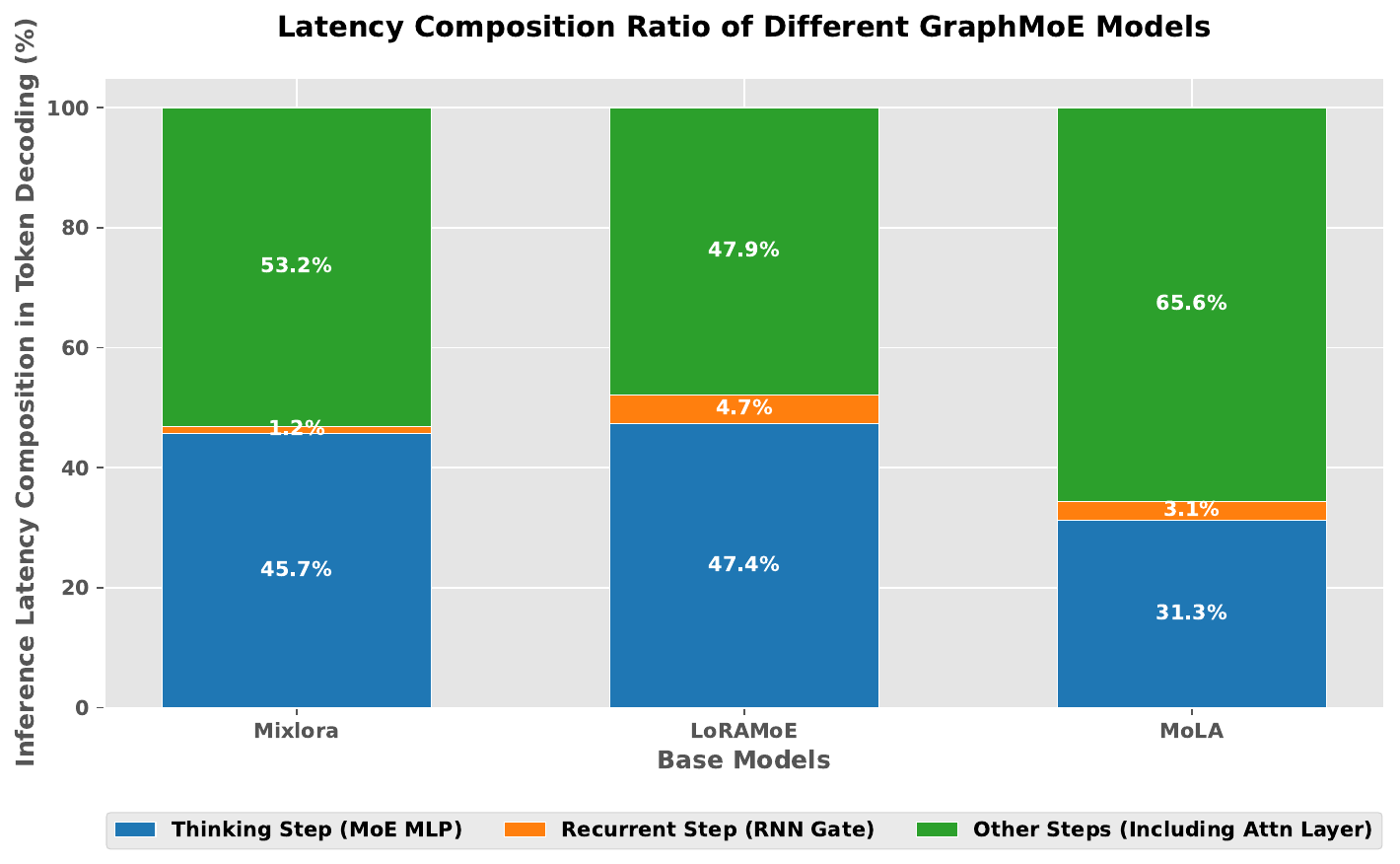}
\vspace{-1em}
\caption{Comparison of the latency compositiong ratio of different based on different LoRA+MoE and \graphmoe{} architectures.}
\label{fig:computation_cost}
\end{figure}

To provide a comprehensive analysis of the increase in time consumption during each rethinking round, we conducted a token-level inference time test for each module executed in the forward step. The results are presented in \autoref{fig:computation_cost}. It is evident that the additional GRU module introduced to the Router contributes marginal running time, accounting for less than $5\%$. This indicates that the majority of increased latency for each additional Reasoning Round ( $T$ ) is primarily due to the additional inference step of MLP module. The time ratios of the Attention Module and the MLP Module determine the slope of the lines shown in (c) of \autoref{fig:sensitivity}. Two key points are highlighted to understand the relationship between module time latency and the inference time latency for each Reasoning Round ( $T$ ):

\begin{itemize}
    \item As the input sequence length increases, the time increase for the Attention Module occurs more rapidly than for the MLP Module. This is because the time complexity of the attention mechanism is $O(N^2)$\footnote{$N$ denotes the length of the input sequence.} due to the multiplication of the Query matrix $Q$ and the Key matrix $K$\footnote{For a detailed description, please refer to \citet{vaswani2017attention}.}. The time complexity of the MLP, being a linear layer, is ( $O(N)$ ). However, due to the implementation of the KV Cache, which includes locally stored cache for historical attention key and value tensors of each layer, the actual time complexity of the Attention Module does not increase as mush as double. For a given task, a longer input sequence length results in a relatively smaller increase in the time for Reasoning Round ( $T$ ).
    \item The different architectures of the MoE LoRA base models influence the time ratio observed in \autoref{fig:computation_cost}. As detailed in \citet{li2024mixlora}, LoRAMoE features vanilla attention LoRA layers along with plugged MLP LoRA layers; MoLA includes plugged attention LoRA layers and plugged MLP LoRA-MoE layers; and MixLora comprises fused attention and fused MLP LoRA-MoE layers. Consequently, during inference, the time spent on MLP LoRA layers is highest for LoRAMoE, followed by MixLoRA, and then MoLA.
\end{itemize}

\section{Conclusion}

In conclusion, this study introduces \graphmoe{}, a novel approach that enhances the cognitive depth of language models through the integration of a self-rethinking mechanism in pseudo-graph Mixture-of-Experts (MoE) networks. Unlike traditional MoE models that operate independently, \graphmoe{} facilitates communication among expert nodes, allowing for iterative refinement and enhanced reasoning capabilities. Implemented using Low-Rank Adaptation techniques (LoRA), \graphmoe{} demonstrates significant performance improvements across various benchmark datasets, outperforming existing LoRA \& LoRA+MoE baseline models and achieving state-of-the-art results. 
This work not only highlights the potential of employing graph-based recurrent routing strategies to implicitly increase the cognitive depth of LLM via ``self-rethinking'', but also opens avenues for further exploration in leveraging interconnected expert networks for advanced problem-solving and reasoning tasks in natural language processing.

\section*{Limitations}
Despite the notable performance improvements achieved by the \graphmoe{} framework, there are several limitations to be acknowledged. 

\begin{itemize}
    \item While the \graphmoe{} architecture consistently leads to performance gains across diverse tasks and base models, the degree of improvement is variable and can depend on the interaction between the MoE and LoRA components. This variability underscores the need for comprehensive exploration into the integration modalities and parameter space to ensure maximum efficiency across different LoRA+MoE configurations.

    \item The workload distribution analysis indicates a marked improvement in workload balance with \graphmoe{}. However, the potential impact of workload imbalance on model performance over a broader range of tasks remains underexplored. Further investigation into balancing expert model selection and activation across diverse scenarios could unveil additional pathways for performance optimization.

    \item The sensitivity analysis indicated potential overfitting issues when increasing the reasoning rounds beyond a particular threshold, revealing the necessity for careful hyperparameter tuning to mitigate issues of over-complexity and model overthinking.

\end{itemize}

In summary, while the \graphmoe{} framework shows promise in enhancing cognitive depth and performance of MoE architectures, future research should address computational precision limits, explore broader integration strategies, and focus on hyperparameter optimization to further refine and substantiate these initial findings.


\normalem
\bibliography{custom}
\bibliographystyle{acl_natbib}

\newpage
\appendix

\section{Discussions}
\label{apx:discussion}
\begin{itemize}
    \item \textbf{What is the value of $k$ in the top-$k$ selection? How is it selected?} Currently, $k$ is treated as a hyperparameter, with two main considerations. First, aligning with other MoE architectures, the number of selected experts typically maintains fixed and sparse, with 2 out of 8 being a common configuration. Second, our framework facilitates expert collaboration, so the top $k$ is not a critical parameter for fully leveraging expert potential.
    \item \textbf{Is the top-k selection in the routing operation (\autoref{equ:topk}) not differentiable?} Regarding the differentiability issue of the top-$k$ in \autoref{equ:topk}, it is important to note that while the top-$k$ operation itself is non-differentiable, it does not directly impact the gradient calculations. Rather, the fixed number of experts chosen through the top-k operation contribute to the subsequent gradient computations.
    \item \textbf{Are some experts learning to do the initial processing, while others more frequently do the later processing?} In this study, the MoE is applied within each transformer block, so the experts do not possess specific physical significance. Therefore, we have not conducted an analysis of individual expert usage patterns or preferences within the reasoning process.
\end{itemize}

\end{document}